Ari Goodman, Glenn Shevach, Sean Zabriskie, Dr. Chris Thajudeen# Cable Slack Detection for Arresting Gear Application using Machine Vision

## ABSTRACT

The cable-based arrestment systems are integral to the launch and recovery of aircraft onboard carriers and on expeditionary land-based installations. These modern arrestment systems rely on various mechanisms to absorb energy from an aircraft during an arrestment cycle to bring the aircraft to a full stop. One of the primary components of this system is the cable interface to the engine. The formation of slack in the cable at this interface can result in reduced efficiency and drives maintenance efforts to remove the slack prior to continued operations.

In this paper, a machine vision based slack detection system is presented. A situational awareness camera is utilized to collect video data of the cable interface region, machine vision algorithms are applied to reduce noise, remove background clutter, focus on regions of interest, and detect changes in the image representative of slack formations. Some algorithms employed in this system include bilateral image filters, least squares polynomial fit, Canny Edge Detection, K-Means clustering, Gaussian Mixture-based Background/Foreground Segmentation for background subtraction, Hough Circle Transforms, and Hough line Transforms. The resulting detections are filtered and highlighted to create an indication to the shipboard operator of the presence of slack and a need for a maintenance action. A user interface was designed to provide operators with an easy method to redefine regions of interest and adjust the methods to specific locations. The algorithms were validated on shipboard footage and were able to accurately identify slack with minimal false positives.## INTRODUCTION

The cable-based arrestment systems are integral to the launch and recovery of aircraft onboard carriers and on expeditionary land-based installations. These modern arrestment systems rely on various mechanisms to absorb energy from an aircraft during an arrestment cycle to bring the aircraft to a full stop. The Advanced Arresting Gear (AAG) is the latest cable based arresting engine utilized by the US NAVY onboard the USS Gerald R. Ford (CVN 78) [1-4]. The system utilizes a cable routing system and a tapered drum to interface the purchase cable with the arresting engine as shown in Figure 1.

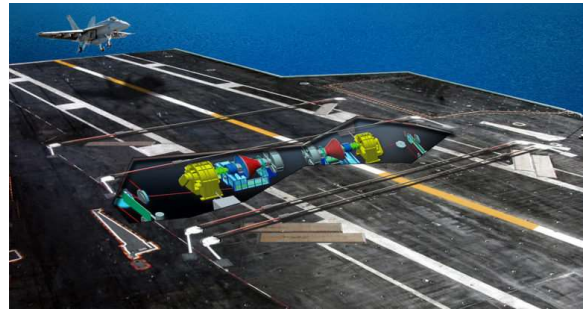

*Figure 1: Advanced Arresting Gear Diagram [3]*

The tension of the cable on the system can be affected by many factors including with weight of the aircraft being arrested and the system presets. When the cable is being wound around the Purchase Cable Drum (PCD), a reduction in tension can lead to the formation of slack. The ability to detect PCD slack is vital to drive maintenance efforts and avoid a potential system availability delay. The current method to detect a slack condition on the PCD is through direct observation by a sailor or engineer. A slack condition can occur during reset of the arresting gear, thus without constant observation there is no way to determine if the slack forms and thus no way to predict if a failure will happen.

NAVAIR Public Release 2023-31 Distribution Statement A - "Approved for public release; distribution is unlimited"

In this work, a machine vision system is presented which provides situational awareness for detecting slack formations on the PCD. The system utilizes existing situational awareness cameras and a-priori knowledge of the PCD physical layout along with machine vision techniques to detect slack formations through change detection techniques. This method employs opensource machine vision algorithms and a count and threshold application to present an indicator to the operator of potential slack formations. The system was validated on shipboard footage.

## METHODOLOGY

The application of machine vision to the detection of slack on the PCD was first undertaken by evaluating various MV approaches to determine slack in a given image using available algorithms from the OpenCV library [5]. Among the approaches evaluated was the utilization of changes at the edges of image components to determine slack formation characteristics. This employed bilateral image filters, least squares polynomial fit, Canny Edge Detection. Clustering algorithms and segmentation algorithms were also investigated, such as the K-Means clustering, and Gaussian Mixture-based Background/Foreground Segmentation for background subtraction. Although these proven techniques all offered benefits and drawbacks, a simplified approach was found to be the best for PCD slack detection. Specifically, exploiting the image angles offered by the existing situational awareness (SA) cameras and physical slack formation geometries a background subtraction and sliding average method with pixel change count thresholding provided a slack detection approach suitable for PCD application.

The situational cameras, utilized for observation of the PCD, output greyscale images from multiple angles within the housing of the PCD. Based on the available interrogation locations for this study, a lower fixed up-facing camera angle was chosen which interrogates the PCD drum, cable, and sheave from the lower right corner. This angle, as shown in the example of Figure 2, offers the most unobstructed view of the PCD and cable for the duration of an arrestment.

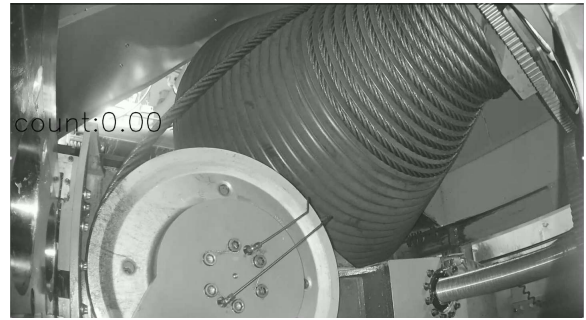

*Figure 2: PCD Initial No Slack Condition*

Although this view offered the most visibility for the regions where slack is likely to occur, the number of moving objects in this view is the highest among possible interrogation angles. As such, in order to employ a change detection methodology, a region of interest was chosen to minimize or eliminate movement from objects not representative of cable slack. The slack region of interest (SRoI) for the selected view is shown in Figure 3 in the checked region of the lower right quadrant of the image. The specific shape of the SRoI is dependent on the specific situational camera angle and view and is shown here for reference only. The final implementation of the algorithm allows each case to have tuned SRoI to maximize detections and reduce false positive detections.

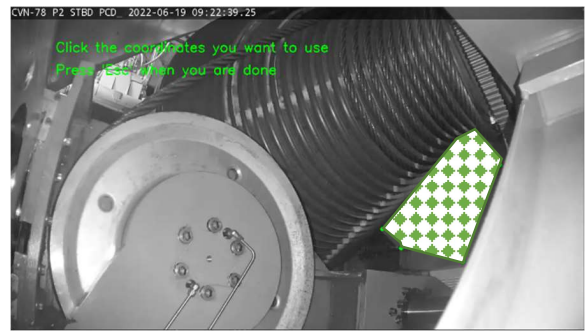

*Figure 3: PCD Slack Region of Interest*

For a given frame of the input video, the machine vision algorithm followed the flow outlined in Figure 4. The image was loaded and the SRoI was chosen. The region was blurred was then blurred followed by background subtraction between the current (N) and previous (N-m) frame of the video. This step eliminated static components within the image. Each image set then applied change detection to determine the pixels in the image that changed between looks. A count is then tabulated for the number of changed pixels. A running average filter is applied to create a Score.

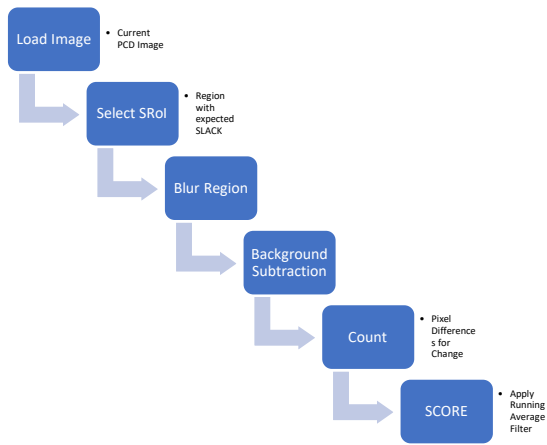

*Figure 4: Machine Vision Process Flow*

## RESULTS

The previously described algorithm for slack detection was evaluated for a series of control images from the Runway Arrested Landing Site (RALS) at JBMDL Lakehurst, which enables in-depth system testing to ensure AAG meets fleet requirements [4]. Utilizing a selected image set from the SA camera, the resulting change events were investigated and thresholds were applied to indicate time stamps associated with slack events of interest. The software system then utilized the time stamps to apply pixel highlighting to the images to give a visual representation of slack formation as an overlay on the output image. For the results discussed, the slack pixels are overlayed in Red on the result images for ease of identification.

The algorithm was evaluated on video feeds from RALS and from arrestment videos provided from the CVN-78 from the existing SA cameras. The images provided were from the PCD SA camera over a 24-hour period during normal operations. Additional videos were provided by the IPT and instrumentation team for historically recorded slack events. These videos were labeled by the Subject Matter Experts (SMEs) to indicate slack, level of slack, and the resulting maintenance actions required.

Due to the lighting and ambient reflections within the PCD enclosure, noise reduction was necessary to reduce the false alarm rates due to background noise in the image showing up as changes between images. An example of typical background noise in these images is shown in Figure 5. To highlight the need for Noise Reduction, the algorithm count was processed and shown in Figure 6. Although the selected video had no slack for the duration of the processed segment, the peaks in Figure 6 indicate slack at the early frames of the image with increases false slack events identified near the middle of the frame set.

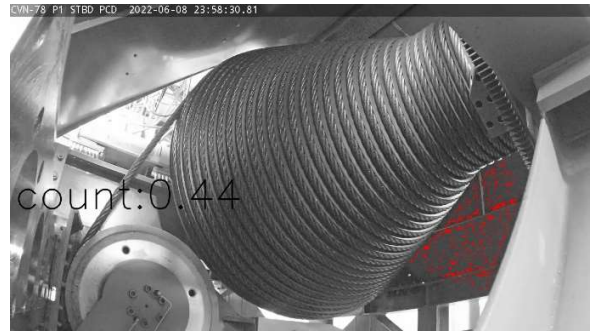

*Figure 5: Noise reduction on PCD Drum Image - CVN-78*

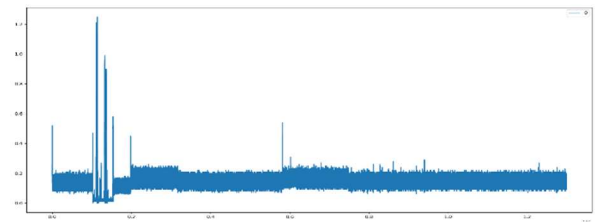

*Figure 6: Count before Noise Reduction*

After employing noise reduction, the videos from the CVN-78 were processed using the completed algorithm. For each of the subsequent results, the change detection was performed between adjacent frames of the video for the first N=0:100 frames. Due to the slow rate of change noted between frames, all frames N>100 were compared to the N-100 frame.

The algorithms employed were able to detect changes in the images that indicate slack. The results from application of the method to hundreds of videos can best be illustrated through representative results from the PCD SA cameras onboard CVN-78 for medium slack conditions. These results indicate the validity of the approach and show validated estimations. Figure 7 shows the detection of slack at three instances across a selected representative arrestment event. The top image in the set shows a small amount of slack occurring early in the arrestment. The indication increases for the same region for the middle and final images for a case where the slack is maintained but not propagated along the PCD for the arrestment.

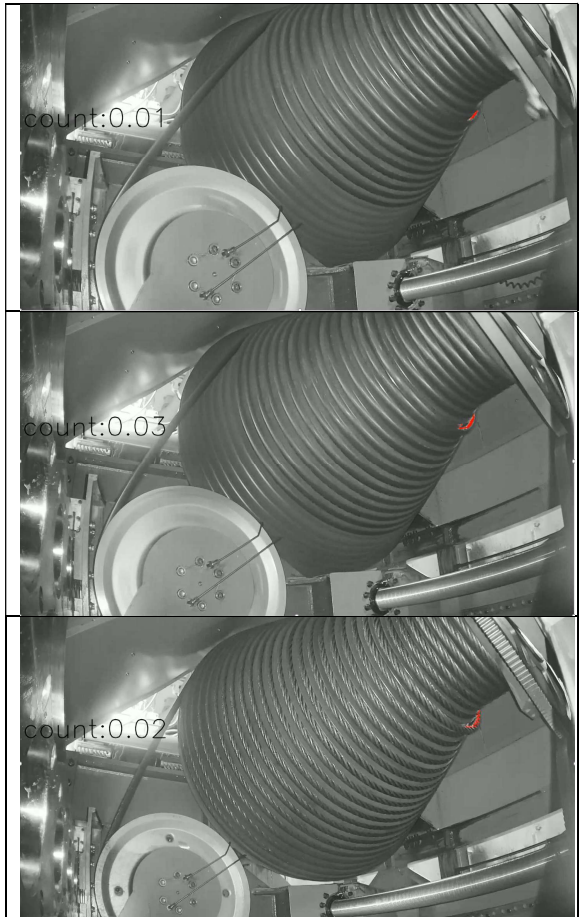

*Figure 7: Machine Vision Slack Detection Across Full Arrestment*

Of particular interest for detection of slack, are those events with "severe" slack detection that propagates along the PCD. Figure 8 and Figure 9 show a case of severe propagating slack with the associated count indicating slack formation. The three visible clusters of peaks in the count can be traced to the formation of slack events that move along the PCD and a more severe compounded slack event at the end of the arrestment indicated by the tall peaks near the end of Figure 9.

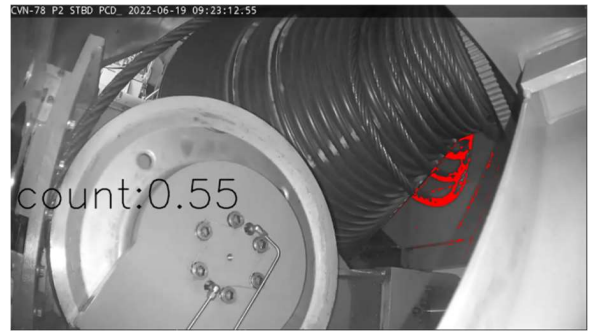

*Figure 8: Extensive Slack detected on PCD - CVN-78*

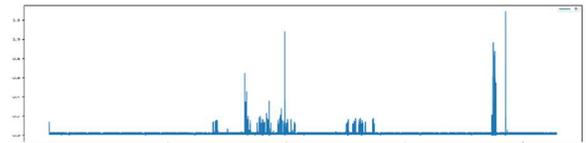

*Figure 9: Count indicating slack formation*

## CONCLUSION

The ability to detect purchase cable drum slack is vital to drive maintenance efforts and avoid a downtime of the system. Currently, a manual inspection is performed using SA cameras with no automated method to detect a slack condition on the PCD without direct observation by a sailor or engineer. A slack condition can occur during reset of the arresting gear, thus without constant observation there is no way to determine if the slack forms and thus no way to predict if a failure will happen. Due to the cable dynamics and forces associated with an arrestment, a slack condition of the purchase cable on the PCD can result in extensive maintenance efforts and suspension of the arresting engine. This would cause the loss of an arresting engine until such time that repairs are made thus reducing the availability and directly affecting operations.

The machine vision algorithm developed leverages computer vision functions available through open-source repositories to determine when slack was forming on the PCD. The algorithm utilizes a SRoI approach with identifies a region of interest where slack is most likely to occur. This region is customizable for applications to existing SA cameras. The algorithms apply noise reduction, blurring, background subtraction, and change averaging methods from Open CV to identify pixels of interest which represent slack formations in a given frame.

The thresholds for detection were tuned and areas of interest were set in order to reduce false positives. The areas of interest were set and the region set were used to determine the number of change pixels for enhanced detection. The change detection procedure uses the current frame compared to the average of previous images.

The finalized algorithms were applied to videos of the PCD during operations onboard CVN-78 to determine the thresholds for operational use. The threshold of detection metrics were analyzed to optimize the detections and reduce false alarm rates. Examples of average non-propagating slack as well as extreme cases of "severe" slack cases on the PCD were presented. The count charts relating slack to pixel change counts were also discussed to show indication of slack.

Future work is planned to adapt this approach to other aspects of the cable arrestment engines for Aircraft Launch and Recovery.

## ACKNOWLEGEMENTS

The authors would like to acknowledge Tyler Comisky, and Alison West for their efforts in designing and evaluating the image processing algorithms and data structures. The authors would also like to acknowledge the Lakehurst Instrumentation team for supporting shipboard data collection.

## AUTHOR BIOS

Ari Goodman is the S&T AI Lead and a Robotics Engineer in the Robotics and Intelligent Systems Engineering (RISE) lab at Naval Air Warfare Center Aircraft Division (NAWCAD) Lakehurst. In this role he leads efforts in Machine Learning, Computer Vision, and Verification & Validation of Autonomous Systems. He received his MS in Robotics Engineering from Worcester Polytechnic Institute in 2017.

Glenn Shevach is the Technical Lead in Diagnostics, Prognostics and Health Management at NAWCAD Lakehurst. He is the Integrated Diagnostics and Automated Test Systems (IDATS) Lab Lead and has led multiple team projects developing PHM technologies for the Navy's Aircraft Launch and Recovery Equipment (ALRE) and Support Equipment (SE). He received his M. Eng. EE from Stevens Institute of Technology in 2010.

Sean Zabriskie is a mechanical engineer and the Electromagnetic Aircraft Launch System (EMALS) and Advanced Arresting Gear (AAG) S&T Portfolio Manager at Naval Air Warfare Center Aircraft Division (NAWCAD) Lakehurst. In this role, he identifies capability gaps, secures funding, and leads research and development efforts related to Ford-class carrier launch and recovery systems. He received his master's in Mechanical Engineering from Stevens Institute of Technology in 2018.

Dr. Chris Thajudeen is the S&T Lead for Advanced Sensing and the Lab Lead Engineer for the Innovative Naval Sensing and Perception Improvement through Research and Engineering (INSPIRE) lab at Naval Air Warfare Center Aircraft Division (NAWCAD) Lakehurst. In this role he leads efforts in Advanced Sensing, Remote Sensing, Sensor Augmentation utilizing Machine Learning, and Electromagnetic Modernization Efforts for Support Equipment (SE) and Aircraft Launch and Recovery Equipment (ALRE). He received his Ph.D. in Electrical Engineering from Villanova University in 2013.